\documentclass[runningheads,a4paper]{llncs}

\usepackage{amssymb}
\usepackage{graphicx}
\usepackage{multirow}
\usepackage{subfigure}
\usepackage{amsmath}

\usepackage{url}
\urldef{\mailsa}\path|{jie_yang, zhiyang_teng}@mymail.sutd.edu.sg |
\urldef{\mailsb}\path| {meishan_zhang, yue_zhang}@sutd.edu.sg |    
\newcommand{\keywords}[1]{\par\addvspace\baselineskip
\noindent\keywordname\enspace\ignorespaces#1}

\begin{document}

\mainmatter  

\title{Combining Discrete and Neural Features for Sequence Labeling}


%
%
\author{Jie Yang \and Zhiyang Teng \and Meishan Zhang \and Yue Zhang\thanks{Corresponding author}}
\authorrunning{Combining Discrete and Neural Features for Sequence Labeling}

\institute{Singapore University of Technology and Design\\
\mailsa\\
\mailsb\\}

%
%

\maketitle

\begin{abstract}
Neural network models have recently\footnote{This paper was accepted by International Conference on Intelligent Text Processing and Computational Linguistics (CICLing) 2016, April; Konya, Turkey.} received heated research attention in the natural language processing community. Compared with traditional models with discrete features, neural models have two main advantages. First, they take low-dimensional, real-valued embedding vectors as inputs, which can be trained over large raw data, thereby addressing the issue of feature sparsity in discrete models. Second, deep neural networks can be used to automatically combine input features, and including non-local features that capture semantic patterns that cannot be expressed using discrete indicator features. As a result, neural network models have achieved competitive accuracies compared with the best discrete models for a range of NLP tasks.

On the other hand, manual feature templates have been carefully investigated for most NLP tasks over decades and typically cover the most useful indicator pattern for solving the problems. Such information can be complementary the features automatically induced from neural networks, and therefore combining discrete and neural features can potentially lead to better accuracy compared with models that leverage discrete or neural features only.

In this paper, we systematically investigate the effect of discrete and neural feature combination for a range of fundamental NLP tasks based on sequence labeling, including word segmentation, POS tagging and named entity recognition for Chinese and English, respectively. Our results on standard benchmarks show that state-of-the-art neural models can give accuracies comparable to the best discrete models in the literature for most tasks and combing discrete and neural features unanimously yield better results.

\end{abstract}

\section{Introduction}

There has been a surge of interest in neural methods for natural language processing over the past few years. Neural models have been explored for a wide range of tasks, including parsing \cite{socher2011parsing,chen2014fast,weiss2015structured,dyer2015transition,zhou2015neural,durrettneural,ballesteros2015transition}, machine translation \cite{kalchbrenner2013recurrent,cho2014learning,sutskever2014sequence,bahdanau2015neural,ling2015character,jean2015using}, sentiment analysis \cite{socher2013recursive,tang2014learning,dos2014deep,vo2015target,zhang2015neural} and information extraction \cite{socher2013reasoning,wang2013effect,ding2015deep,mark2016what}, achieving results competitive to the best discrete models.

Compared with discrete models with manual indicator features, the main advantage of neural networks is two-fold. 
First, neural network models take low-dimensional dense embeddings \cite{mikolov2013efficient,pennington2014glove,collobert2011natural} as inputs, which can be trained from large-scale test, thereby overcoming the issue of sparsity. 
Second, non-linear neural layers can be used for combining features automatically, which saves the expense of feature engineering. 
The resulting neural features can capture complex non-local syntactic and semantic information, which discrete indicator features can hardly encode.

On the other hand, discrete manual features have been studied over decades for many NLP tasks, and effective feature templates have been well-established for them. 
This source of information can be complementary to automatic neural features, and therefore a combination of the two feature sources can led to improved accuracies. 
In fact, some previous work has attempted on the combination. 
Turian et al. \cite{turian2010word} integrated word embedding as real-word features into a discrete Conditional Random Field \cite{crf} (CRF) model, finding enhanced results for a number of sequence labeling tasks. 
Guo et al. \cite{guo2014revisiting} show that the integration can be improved if the embedding features are carefully discretized. 
On the reverse direction, Ma et al. \cite{ma2014tagging} treated a discrete perception model as a neural layer, which is integrated into a neural model.
Wang \& Manning \cite{wang2013learning} integrated a discrete CRF model and a neural CRF model by combining their output layers.   
Greg \& Dan \cite{durrettneural} and Zhang et al. \cite{zhang2015neural} also followed this method. 
Zhang \& Zhang \cite{zhangcombining} compared various integration methods for parsing, and found that the second type of integration gives better results.

We follow Zhang \& Zhang \cite{zhangcombining}, investigating the effect of feature combination for a range of sequence labeling tasks, including word segmentation, 
Part-Of-Speech (POS) tagging and named entity recognition (NER) for Chinese and English, respectively. 
For discrete features, we adopt a CRF model with state-of-the art features for each specific task. 
For neural features, we adopt a neural CRF model, using a separated Long Short-Term Memory \cite{hochreiter1997long} (LSTM) layer to extract input features. 
We take standard benchmark datasets for each task. 
For all the tasks, both the discrete model and the neural model give accuracies that are comparable to the state-of-the-art. 
A combination of discrete and neural feature, gives significantly improved results with no exception. 

The main contributions that we make in this investigation include:

- We systematically investigate the effect of discrete and neural feature combination for a range of fundamental NLP tasks, showing that the two types of feature are complimentary.

- We systematically report results of a state-of-the-art neural network sequence labeling model on the NLP tasks, which can be useful as reference to future work.

- We report the best results in the literatures for a number of classic NLP tasks by exploiting neural feature integration.

- The source code of the LSTM and CRF implementations of this paper are released under GPL at https://github.com/SUTDNLP/NNSegmentation, ../NNPOSTagging and ../NNNamedEntity .

\section{Related Work}
There has been two main kinds of methods for word segmentation. 
Xue \cite{xue2003chinese} treat it as a sequence labeling task, using B(egin)/I(nternal) /E(nding)/S(ingle-character word) tags on each character in the input to indicate its segmentation status. 
The method was followed by Peng et al. \cite{peng2004chinese}, who use CRF to improve the accuracies. 
Most subsequent work follows  \cite{peng2004chinese,zhao2009character,jiang2008cascaded,sun2011stacked,liu2014domain} and 
feature engineering has been out of the key research questions. 
This kind of research is commonly referred to as the character-based method. 
Recently, neural networks have been applied to character-based segmentation \cite{zheng2013deep,pei2014maxmargin,chen2015gated}, giving results comparable to discrete methods. 
On the other hand, the second kind of work studies word-based segmentation, scoring outputs based on word features directly  \cite{zhang2007chinese,sun2010word,liu2012unsupervised}. 
We focus on the character-based method, which is a typical sequence labeling problem.

POS-tagging has been investigated as a classic sequence labeling problem  \cite{ratnaparkhi1996maximum,collins2002discriminative,manning2011part},
for which a  well-established set of features are used. 
These handcrafted features basically include words, the context of words,  word morphologies and word shapes. 
Various neural network models have also been used for this task. 
In order to include word morphology and word shape knowledge, a convolutional neural network (CNN) for automatically learning character-level representations is investigated in \cite{santos2014learning}.
Collobert et al. \cite{collobert2011natural} built a CNN neural network for multiple sequence labeling tasks, which gives state-of-the-art POS results. 
Recurrent neural network models have also been used for this task \cite{perez2001part,Huang:2015aa}. 
Huang et al. \cite{Huang:2015aa} combines bidirectional LSTM with a CRF layer, their model is robust and has less dependence on word embedding.

Named entity recognition is also a classical sequence labeling task in the NLP community. 
Similar to other tasks, most works access NER problem through feature engineering. 
McCallum \& Li \cite{mccallum2003early} use CRF model for NER task and exploit Web lexicon as feature enhancement.
Chieu \& Ng \cite{hai2003named}, Krishnan \& Manning \cite{krishnan2006effective} and Che et al. \cite{che2013named} tackle this task through non-local features \cite{ratinov2009design}. 
Besides, many neural models, which are free from handcrafted features, have been proposed in recent years. 
In Collobert et al. \cite{collobert2011natural} model we referred before, NER task has also been included. 
Santos et al. \cite{dos2015boosting} boost the neural model by adding character embedding on Collobert's structure. 
James et al. \cite{Hammerton:2003:CONLL} take the lead by employing LSTM for NER tasks. 
Chiu et al. \cite{chiu2015named} use CNN model to extract character embedding and attach it with word embedding and afterwards feed them into Bi-directional LSTM model. 
Through adding lexicon features, Chiu's NER system get state-of-the-art performance on both CoNLL2003 and OntoNotes 5.0 NER datasets.

\section{Method}

\begin{figure}[!t] 
  \centering 
  \subfigure[Discrete Model Structure]{ 
    \label{fig:crf} 
    \includegraphics[width=4.0in]{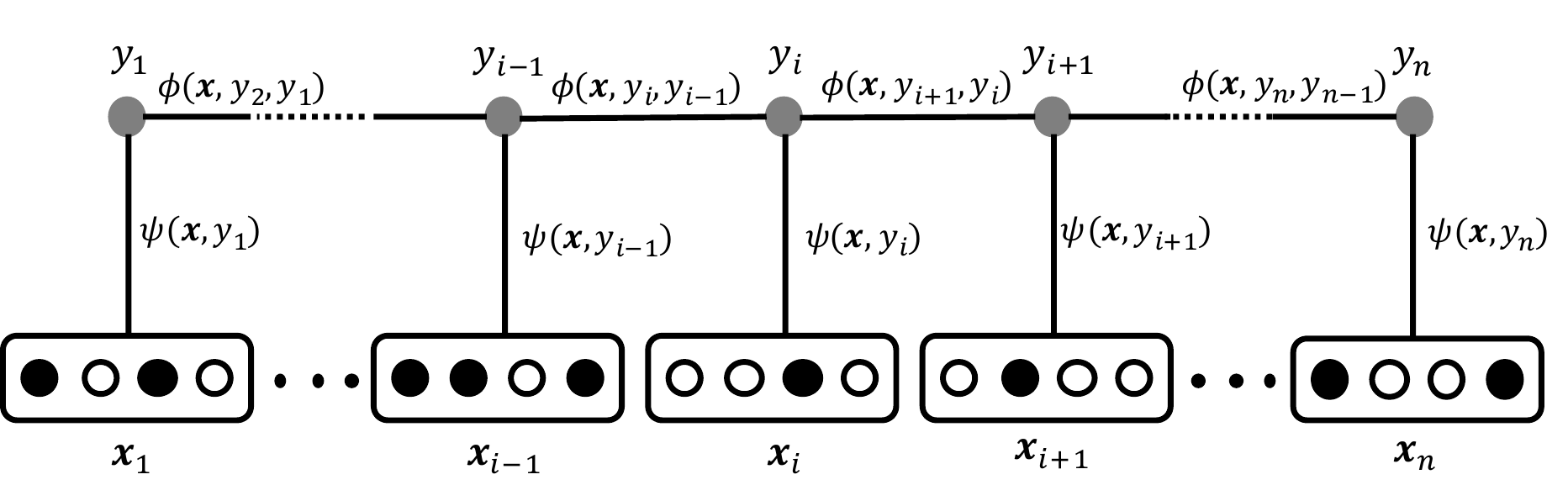}} 
  \subfigure[Neural Model Structure]{ 
    \label{fig:neural} 
    \includegraphics[width=4.0in]{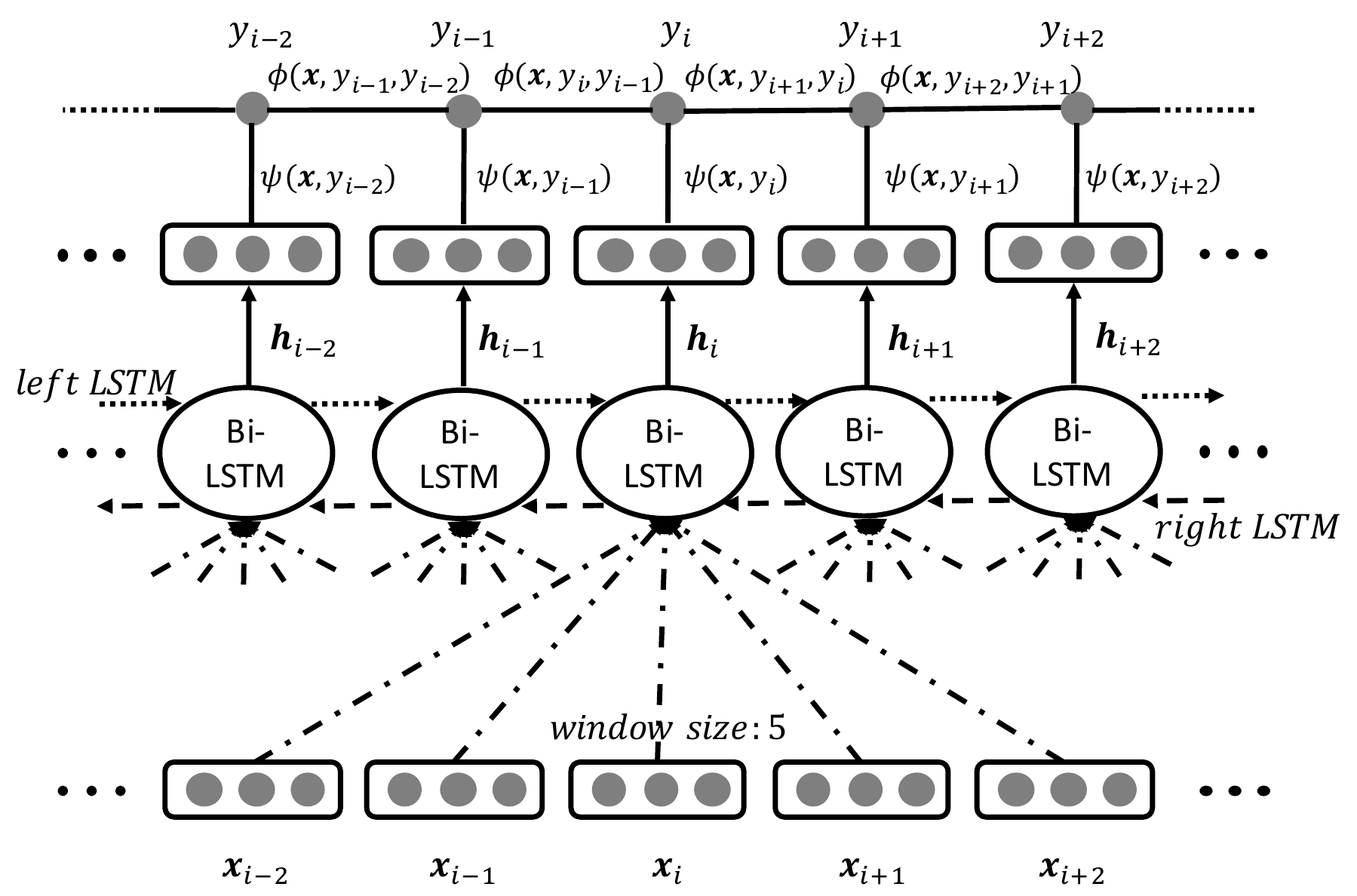}} 
  \caption{Model Structures} 
  \label{models} 
\end{figure}

The structures of our discrete and neural models are shown in Fig. \ref{fig:crf} and \ref{fig:neural},  respectively, which are used for all the tasks in this paper. Black and white elements represent binary features for discrete model and gray elements are continuous representation for word/character embedding.
The only difference between different tasks are the definition of input and out sequences, and the features used. 

\textbf{The discrete model} is a standard CRF model.  Given a sequence of input $\vec{x} = x_1, x_2, \dots, x_n$, it models the output sequence $\vec{y} = y_1, y_2, \dots, y_n$
by calculating two potentials.  
In particular, the \emph{output clique potential} shows the correlation between inputs and output labels, 
\begin{equation}
    \Psi(\vec{x}, y_i) = \exp (\vec{\theta_o} \cdot \vec{f_o} ( \vec{x}, y_i) )
\end{equation}
where $\vec{f_o} ( \vec{x}, y_i) $ is a feature vector extracted from $\vec{x}$ and $y_i$, and $\vec{\theta_o}$ is a parameter vector. 

The \emph{edge clique potential} shows the correlation between consecutive output labels, 
\begin{equation}
    \Phi( \vec{x}, y_i, y_{i-1}) = \exp ( \vec{\theta_e} \cdot \vec{f_e}(\vec{x}, y_i, y_{i-1}))
\end{equation}
where $\vec{f_e}(\vec{x}, y_i, y_{i-1})$ is a feature vector extracted from $\vec{x}$, $y_i$ and $y_{i-1}$, and $\vec{\theta_e}$ is a parameter vector. 

The final probability of $\vec{y}$ is estimated as 
\begin{equation}
 p(\vec{y} | \vec{x} ) = \frac{\prod_{i=1}^{|\vec{y}|} \Psi(\vec{x}, y_i)  \prod_{j=1}^{|\vec{y}|} \Phi(\vec{x}, y_j, y_{j-1}) }{Z(\vec{x})}\end{equation}
where $Z(\vec{x})$ is the partition function, 
\begin{equation} 
   Z(\vec{x}) = \sum_{\vec{y}}\prod_{i=1}^{|\vec{y}|} \Psi(\vec{x}, y_i)  \prod_{j=1}^{|\vec{y}|} \Phi(\vec{x}, y_j, y_{j-1})  
\end{equation}

The overall features $\{\vec{f_o} ( \vec{x}, y_i) $, $\vec{f_e}(\vec{x}, y_i, y_{i-1})\}$  are extracted at each location $i$ according to a set of feature templates for each task. 

\textbf{ The neural model} takes the neural CRF structure. 
Compared with the discrete model, it replaces the output clique features $ \vec{f_o} ( \vec{x}, y_i) )$ with a dense neural feature vector $\vec{h_i}$, which is computed using neural network layer,
\begin{equation}
    \begin{split}
        \vec{h_i} = BiLSTM ((\vec{e}(x_{i-2}), \vec{e}(x_{i-1}),& \vec{e}(x_i), \vec{e}(x_{i+1}), \vec{e}(x_{i+2})), \vec{W}, \vec{b}, \vec{h_{i-1}} ) \\
        \Psi(\vec{x}, y_i) &= \exp(\vec{\theta_o} \cdot \vec{h_i} )\\
    \end{split}
\end{equation}
where $\vec{e}(x_i)$ is the embedding form of $x_i$, \emph{BiLSTM} represents bi-directional LSTM structure for calculating hidden state $\vec{h_i}$ for input $x_i$, which considers both the left-to-right and right-to-left information flow in a sequence. The \emph{BiLSTM} structure receives input of the embeddings from a window of size 5 as shown in Fig. \ref{fig:neural}.

For the neural model, the edge clique is replaced with a single transition weight $\tau(y_i, y_{i-1})$.  The remainder of the model is the same as the discrete model 
\begin{equation}
 \begin{split}
   p(\vec{y} | \vec{x} ) &= \frac{\prod_{i=1}^{|\vec{y}|} \Psi(\vec{x}, y_i)  \prod_{j=1}^{|\vec{y}|} \Phi(\vec{x}, y_j, y_{j-1}) }{Z(\vec{x})}\\
	                              &= \frac{\prod_{i=1}^{|\vec{y}|} \exp(\vec{\theta_o}\cdot \vec{h_i})  \prod_{j=1}^{|\vec{y}|} \exp(\tau(y_j, y_{j-1}))  }{Z(\vec{x})}\\
 \end{split}
\end{equation}

Here $\vec{\theta_o}$ and $\tau(y_i, y_{i-1})$ are model parameters, which are different from the discrete model. 

\textbf{The joint model}  makes a concatenation of the discrete and neural features at the output cliques and edge cliques, 
\begin{equation}
\begin{split}
  \Psi(\vec{x}, y_i) &= \exp( \vec{\theta_o} \cdot ( \vec{h_i} \oplus \vec{f_o} ( \vec{x}, y_i) ) ) \\
 \Phi(\vec{x}, y_j, y_{j-1}) &= \exp(\vec{\theta_e}  \cdot ( [\tau(y_j, y_{j-1})] \oplus \vec{f_e}(\vec{x}, y_j, y_{j-1}) )) \\
\end{split}
\end{equation}
where the $\oplus$ operator is the vector concatenation operation. 

\textbf{The training objective} for all the models is to maximize the margin between gold-standard and model prediction scores. 
Given a set of training examples $\{\vec{x_n}, \vec{y_n})\}_{n=1}^N$,  the objective function is defined as follows
\begin{equation}
 L = \frac{1} {N} \sum_{n=1}^N loss(\vec{x_n}, \vec{y_n},  \vec{\Theta}) + \frac{\lambda}{2} ||\vec{\Theta}||^2
\end{equation}

Here  $\vec{\Theta}$ is the set of model parameters $\vec{\theta_o}$, $\vec{\theta_e}$, $\vec{W}$, $\vec{b}$ and $\tau$, and $\lambda$ is the $L_2$ regularization parameter.   

The loss function is defined as 
\begin{equation}
 loss(\vec{x_n}, \vec{y_n},  \vec{\Theta})  = \max_{\vec{y}} ( p( \vec{y} | \vec{x_n}; \vec{\Theta}) + \delta(\vec{y}, \vec{y_n}))- p(\vec{y_n} | \vec{x_n}, \vec{\Theta})
\end{equation}
where $\delta(\vec{y}, \vec{y_n})$ denotes the hamming distance between $\vec{y}$ and $\vec{y_n}$. 

Online Adagrad \cite{duchi2011adaptive} is used to train the model, with the initial learning rate set to be $\eta$. Since the loss function is not differentiable, 
a subgradient is used, which is estimated as 
\begin{equation}
\frac{\partial{ loss (\vec{x_n}, \vec{y_n}, \vec{\Theta}) }}{\partial{\vec{\Theta}}} =  \frac{\partial{p(\vec{\widehat{y} } |\vec{x_n}, \vec{\Theta} )}}{\partial{\vec{\Theta}}} -  \frac{\partial{p(\vec{y_n} |\vec{x_n}, \vec{\Theta} )}}{\partial{\vec{\Theta}}} 
\end{equation}
where $\vec{\widehat{y}}$ is the predicted label sequence. 

\textbf{Chinese Word Segmentation Features.} Table \ref{tab:SegFeature} shows the features used in Chinese word segmentation. For ``type'', each character has five possibilities: 0/Punctuation, 1/Alphabet, 2/Date, 3/Number and 4/others.
\begin{table}[!htbp]
\centering
\caption{Feature Templates for Chinese Word Segmentation}\label{tab:SegFeature}
\begin{tabular}{c|l}
\hline
1 & character unigram: $c_i$ ,$-2\leq i \leq 2$\\
2 & character bigram: $c_{i-1}c_i$ ,$-1\leq i \leq 2$; $c_{-1}c_{1}$ and $c_{0}c_{2}$\\
3 & whether two characters are equal or not: $c_0==c_{-2}$ and $c_0==c_{1}$ \\
4 & character trigram: $c_{-1}c_{0}c_1$\\
5 & character type unigram: $type(c_0)$ \\
6 & character types: $type(c_{-1})type(c_0)type(c_1)$ \\
7 & character types: $type(c_{-2})type(c_{-1})type(c_0)type(c_1)type(c_{2})$ \\
\hline
\end{tabular}
\end{table}

\textbf{POS Tagging Features.} Table \ref{tab:TagFeature} lists features for POS tagging task on both English and Chinese datasets. 
The prefix and suffix include 5 characters for English and 3 characters for Chinese. 
\begin{table}[!htbp]
\centering
\caption{Feature Templates for POS tagging}\label{tab:TagFeature}
\begin{tabular}{c|l}
\hline
1& word unigram: $w_i, -2\le i\le 2$\\
2& word bigram: $w_{-1}w_0, w_0w_1, w_{-1}w_1$\\
3& prefix: $Prefix(w_0)$\\
4& suffix: $Suffix(w_0)$\\
5& length: $Length(w_0)$ (only for Chinese)\\
\hline
\end{tabular}
\end{table}

\textbf{NER Features.} Table \ref{tab:EngNerFeature} shows the feature template used in English NER task. 
For ``word shape'', each character in word is located in one of these four types: number, lower-case English character, upper-case English character and others. 
``Connect'' word has five categories: ``of'', ``and'', ``for'', ``-'' and other. Table \ref{tab:ChnNerFeature} presents the features used in Chinese NER task. 
We extend the features used in Che et al. \cite{che2013named} by adding part-of-speech information. Both POS tag on English and Chinese datasets are labeled by ZPar \cite{zhang2011syntactic}, prefix and suffix on two datasets are both including 4 characters.
Word clusters in both English and Chinese tasks are same with Che's work \cite{che2013named}.

\begin{table}[!t]
\centering
\caption{Feature Templates for English NER}\label{tab:EngNerFeature}
\begin{tabular}{c|l}
\hline
1& word unigram: $w_i, -1\le i\le 1$\\
2& word bigram: $w_iw_{i+1}, -2\le i \le 1$\\
3& word shape unigram: $Shape(w_i), -1 \le i \le 1$\\
4& word shape bigram: $Shape(w_i)Shape(w_{i+1}), -1\le i \le 0$\\
5& word capital unigram: $Capital(w_i), -1 \le i \le 1$\\
6& word capital with word: $Capital(w_i) w_j, -1 \le i, j \le 1 $\\
7& connect word unigram: $Connect(w_i), -1 \le i\le 1$\\
8& capital with connect: $Capital(w_i) Connect(w_0), -1 \le i \le 1$\\
9& word cluster unigram: $Cluster(w_i), -1 \le i \le 1$\\
10& word cluster bigram: $Cluster(w_i)Cluster(w_{i+1}, -1\le i \le 0)$\\
11& word prefix: $Prefix(w_i), 0 \le i \le 1$\\
12& word suffix: $Suffix(w_i), -1 \le i \le 0$\\
13& word POS unigram: $POS(w_0)$\\
14& word POS bigram: $POS(w_i)POS(w_{i+1}), -1 \le i \le 0$\\
15& word POS trigram: $POS(w_{-1})POS(w_0)POS(w_1)$\\
16& POS with word: $POS(w_0)w_0$\\
\hline
\end{tabular}
\end{table}

\begin{table}[!t]
\centering
\caption{Feature Templates for Chinese NER}\label{tab:ChnNerFeature}
\begin{tabular}{c|l}
\hline
1& word POS unigram: $POS(w_0)$\\
2& word POS bigram: $POS(w_i)POS(w_{i+1}), -1 \le i \le 0$\\
3& word POS trigram: $POS(w_{-1})POS(w_0)POS(w_1)$\\
4& POS with word: $POS(w_0)w_0$\\
6& word unigram: $w_{i}, -1\le i \le 1$\\
7& word bigram: $w_{i-1}w_{i}, 0\le i \le 1$\\
8& word prefix: $Prefix(w_i), -1 \le i \le 0$\\
9& word prefix: $Suffix(w_i), -1 \le i \le 0$\\
10& radical: $radical(w_0, k), 0\le k \le 4$; $k$ is character position \\
11& word cluster unigram: $Cluster(w_0)$\\
\hline
\end{tabular}
\end{table}

\section{Experiments}
We conduct our experiments on different sequence labeling tasks, including Chinese word segmentation, Part-of-speech tagging and Named entity recognition. 

For these three tasks, their input embeddings are different. For Chinese word segmentation, we take both character embeddings and character bigram embeddings for calculating $\vec{e}(x_i)$. For POS tagging, $\vec{e}(x_i)$ consists of word embeddings and  character embeddings. For NER, we include word embeddings, character embeddings and POS embeddings for $\vec{e}(x_i)$.
Character embeddings, character bigram embeddings and word embeddings are pretrained separately using word2vec\cite{mikolov2013efficient}. English word embedding is chosen as SENNA \cite{collobert2011natural}. We make use of Chinese Gigaword Fifth Edition\footnote[1]{https://catalog.ldc.upenn.edu/LDC2011T13} to pretrain necessary embeddings for Chinese words. The Chinese corpus is segmented by ZPar \cite{zhang2011syntactic}.
During training, all these aforementioned embeddings will be fine-tuned. The hyper-parameters in our experiments are shown at Table \ref{tab:hyperparameter}. Dropout \cite{srivastava2014dropout} technology has been used to suppress over-fitting in the input layer. 
\begin{table}[!t]
\centering
\caption{Hyper Parameters}\label{tab:hyperparameter}
\begin{tabular}{l|c}
\hline
Parameter & Value\\
\hline
dropout probability & 0.25\\
wordHiddensize & 100\\
charHiddensize & 60 \\
charEmbSize & 30\\
wordEmbSize & 50 \\
wordEmbFineTune & True\\
charEmbFineTune & True\\
initial $\eta$ & 0.01\\
regularization $\lambda$ & 1e-8\\
\hline
\end{tabular}
\end{table}

\subsection{Chinese Word Segmentation}
For Chinese word segmentation, we choose PKU, MSR and CTB60 as evaluation datasets. 
The PKU and MSR dataset are obtained from SIGHAN Bakeoff 2005 corpus\footnote[3]{http://www.sighan.org/bakeoff2005}. 
We split the PKU and MSR datasets in the same way as Chen et al. \cite{chen-EtAl:2015:EMNLP2}, and the CTB60 set as Zhang et al. \cite{zhang2014character}. Table \ref{tab:segdatasets}
shows the statistical results for these three datasets. We evaluate segmentation accuracy by Precision (P), Recall (R) and F-measure (F). 
\begin{table}[!htbp]
\centering
\caption{Chinese Word Segmentation Datasets Statistics}\label{tab:segdatasets}
\begin{tabular}{|c|c|c|c|}
\hline
Segmentation datasets &\multirow{2}*{PKU} & \multirow{2}*{MSR} &\multirow{2}*{CTB60}\\
(sentences) & & &\\
\hline
Train& 17149& 78226 &23401\\
\hline
Dev& 1905& 8692 & 2078\\
\hline
Test& 1944& 3985 & 2795\\
\hline
\end{tabular}
\end{table}

\begin{figure} 
  \centering 
  \subfigure[CTB]{ 
    \label{fig:ctb} 
    \includegraphics[width=1.5in]{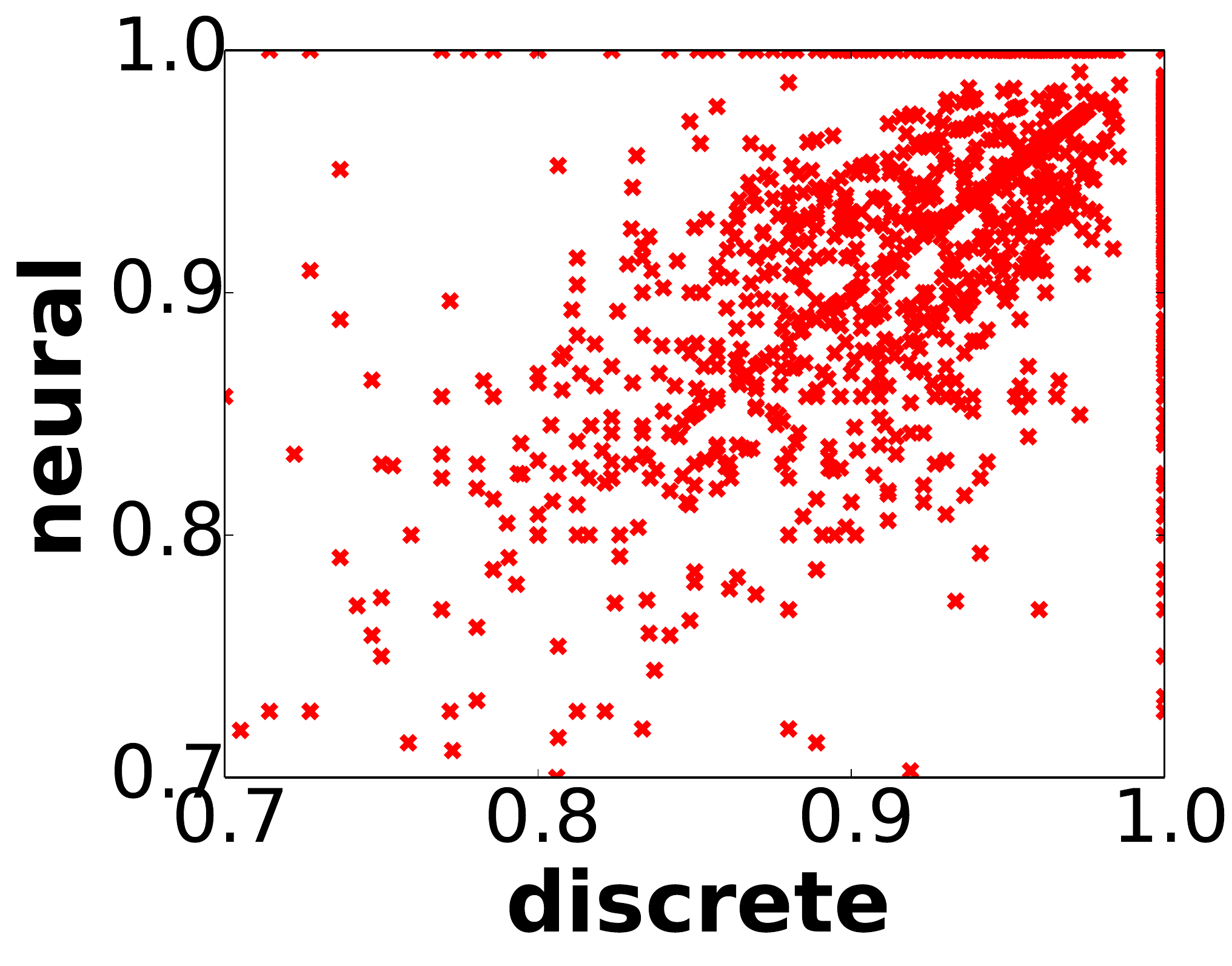}} 
  \subfigure[MSR]{ 
    \label{fig:msr} 
    \includegraphics[width=1.5in]{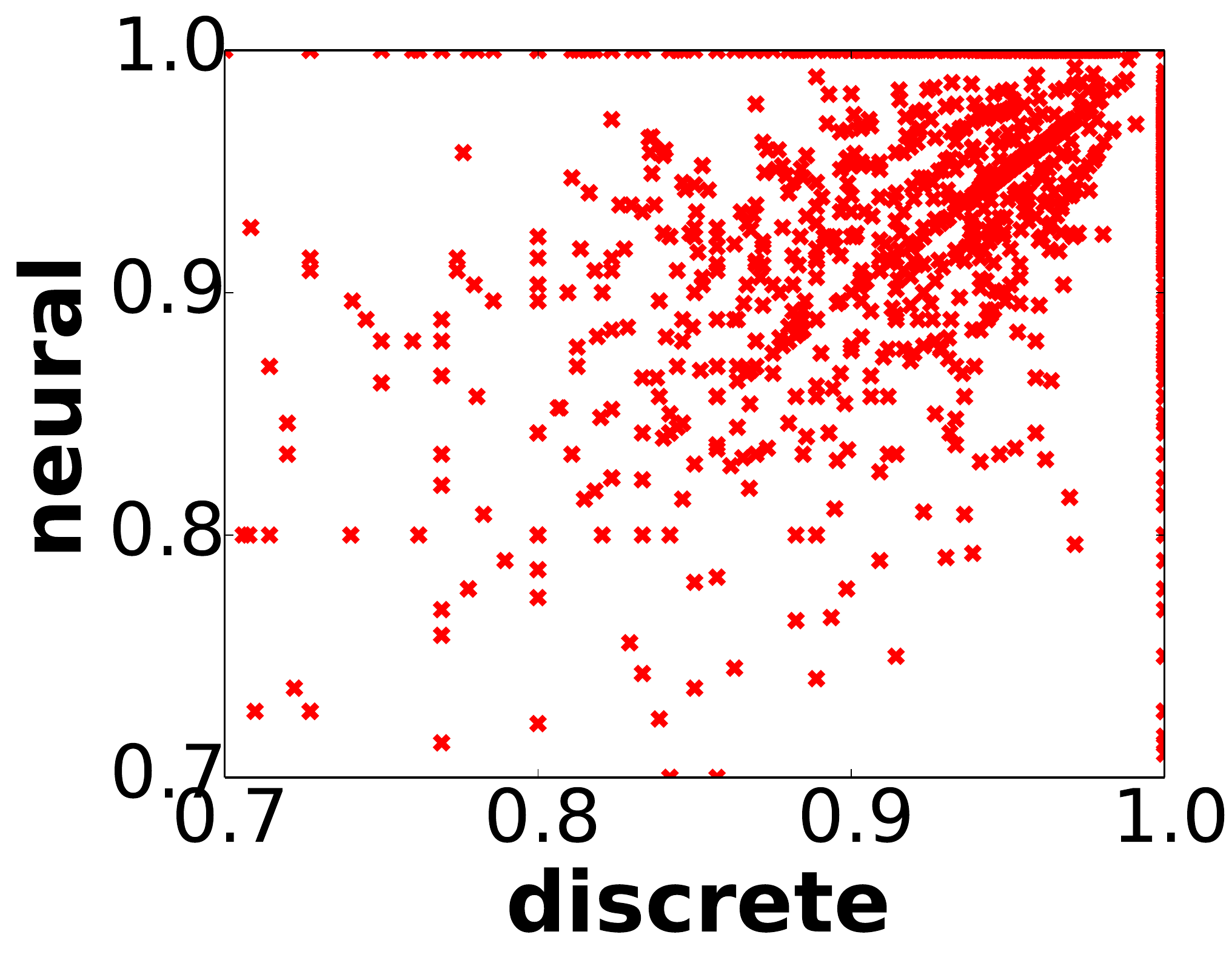}} 
  \subfigure[PKU]{ 
    \label{fig:pku} 
    \includegraphics[width=1.5in]{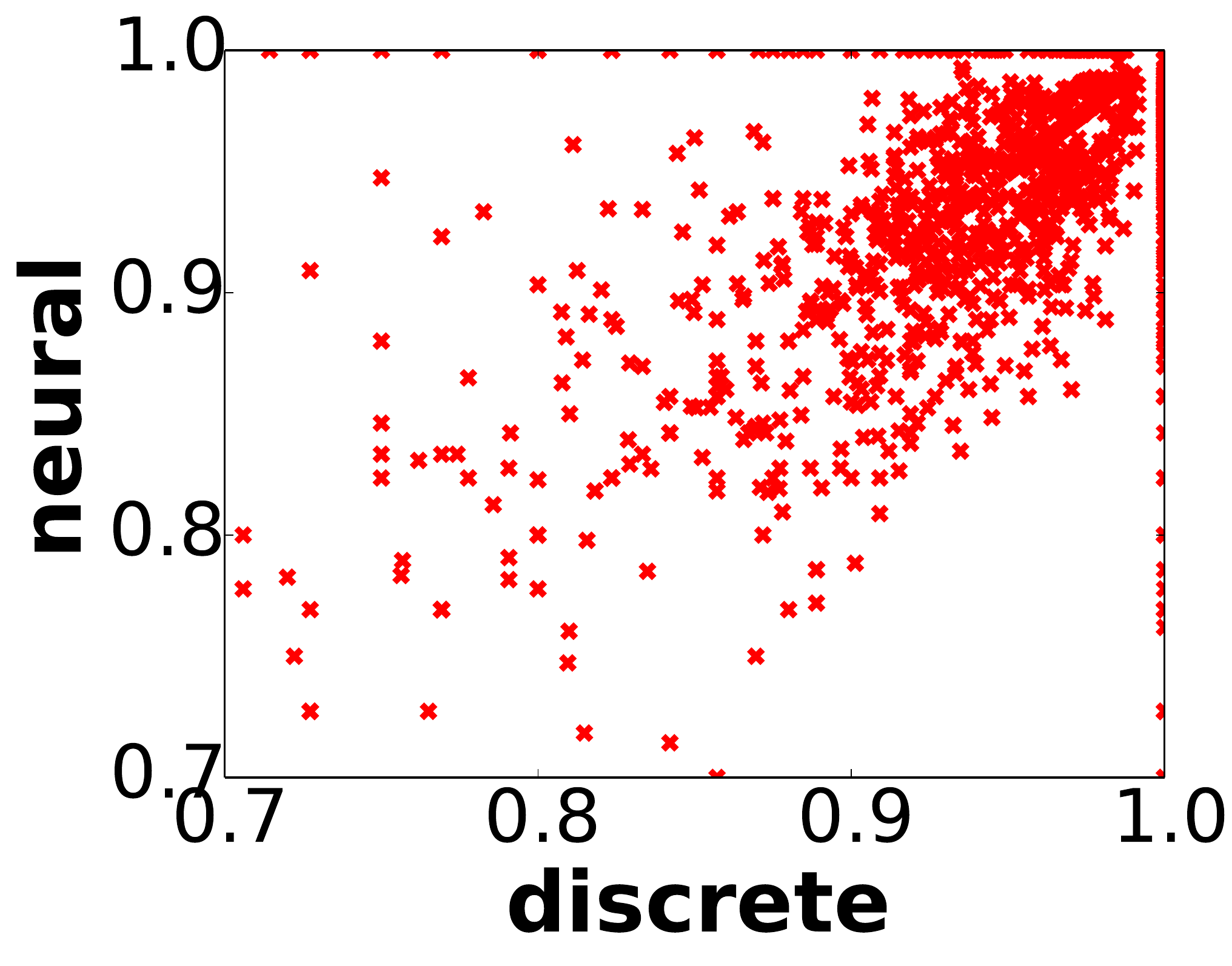}} 
  \caption{Chinese Word Segmentation F-measure Comparisons} 
  \label{segscatter} 
\end{figure}

The experiment results of Chinese word segmentation is shown in Table \ref{tab:wordseg}. Our joint models shown
comparable results to the state-of-the-art results reported by Zhang \& Clark \cite{zhang2007chinese}, where they 
adopt a word-based perceptron model with carefully designed discrete features. Compared with both discrete and neural models, our joint models can achieve best results among all three datasets. In particular, the joint model can 
outperform the baseline discrete model by 0.43, 0.42 and 0.37 on PKU, MSR, CTB60 respectively. In order to investigate
whether the discrete model and neural model can benefit from each other, we scatter sentence-level segmentation accuracy of two models for three datasets in Fig. \ref{segscatter}.  As we can see from Fig. \ref{segscatter}, some sentences can obtain higher accuracies in the neural model, while other sentences can win out in the discrete model. 
This common phenomenon among three datasets suggests that the neural model and the discrete model can be combined together to enjoy the merits from each side.

\begin{table}[!htbp]
\centering
\caption{Chinese Word Segmentation Results}\label{tab:segresults}
\label{tab:wordseg}
\begin{tabular}{|c|ccc|ccc|ccc|}
\hline
\multirow{2}*{Model}& \multicolumn{3}{|c|}{PKU}& \multicolumn{3}{|c|}{MSR}& \multicolumn{3}{|c|}{CTB60}\\
\cline{2-10}
 & P& R& F& P& R& F& P& R& F\\
\hline
Discrete& 95.42& 94.56& 94.99& 96.94& 96.61& 96.78& 95.43& 95.16& 95.29\\
\hline
Neural& 94.29& 94.56& 94.42& 96.79& \textbf{97.54}& 97.17& 94.48& 95.01& 94.75\\
\hline
Joint& \textbf{95.74}& \textbf{95.12}& \textbf{95.42}& \textbf{97.01}& 97.39& \textbf{97.20}& \textbf{95.68}& \textbf{95.64}& \textbf{95.66}\\
\hline
State-of-the-art& N/A& N/A& 94.50& N/A& N/A& \textbf{97.20}& N/A& N/A& 95.05\\
\hline
\end{tabular}
\end{table}

\subsection{POS Tagging}
We compare our models on both English and Chinese datasets for the POS tagging task. The English dataset is chosen following Toutanova et al. \cite{toutanova2003feature} and Chinese dataset by Li et al. \cite{zhenghua2015coupled} on CTB. Statistical results are shown in Table \ref{tab:posdatasets}. Toutanova's model \cite{toutanova2003feature} exploits bidirectional dependency networks to capture both preceding and following tag contexts for English POS tagging task. Li et al. \cite{zhenghua2015coupled} utilize heterogeneous datasets for Chinese POS tagging through bundling two sets of tags and training in enlarged dataset, their system got state-of-the-art accuracy on CTB corpus. 

\begin{figure} 
  \centering 
  \subfigure[English]{ 
    \label{fig:posen} 
    \includegraphics[width=2.2in]{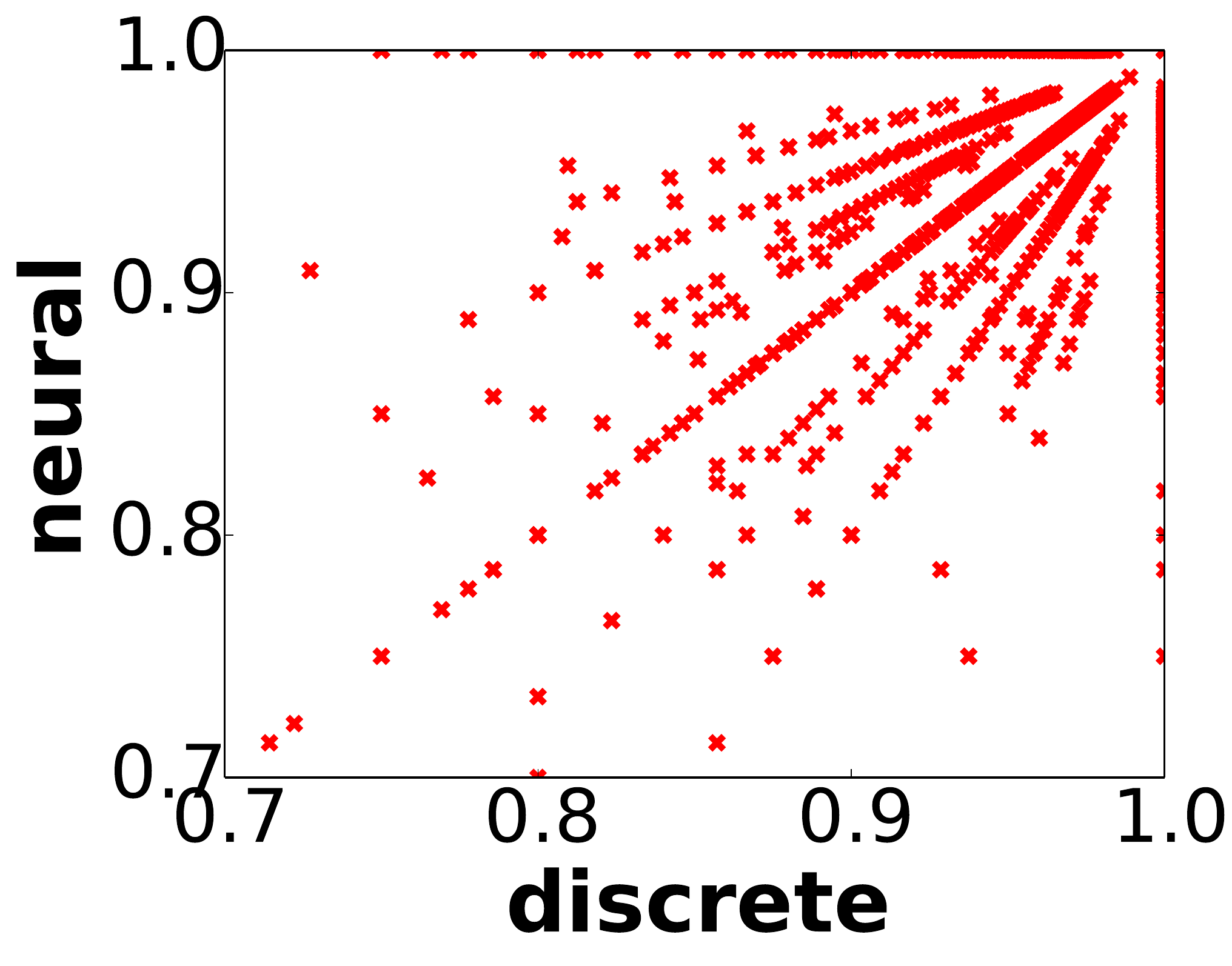}} 
  \subfigure[Chinese]{ 
    \label{fig:poscn} 
    \includegraphics[width=2.2in]{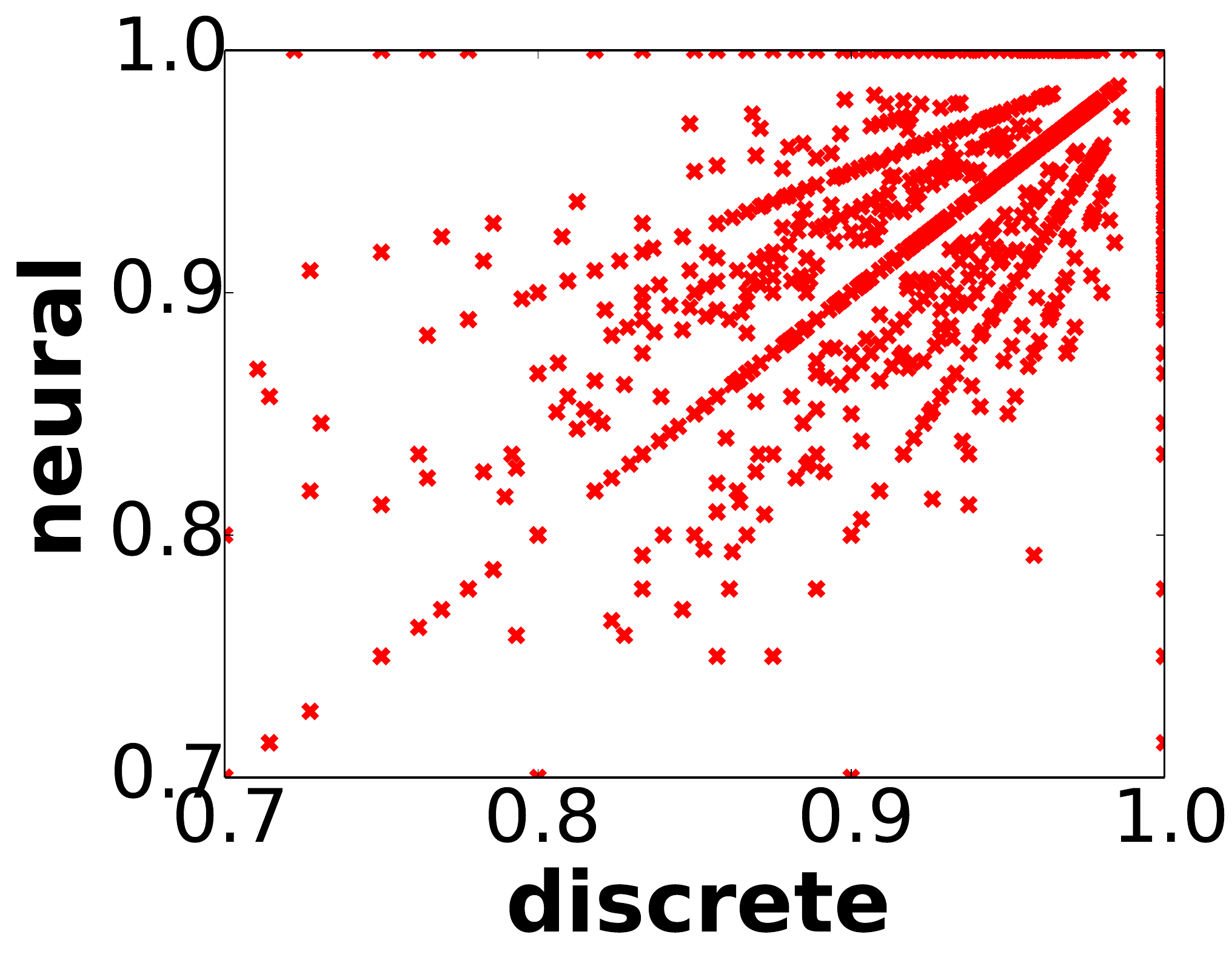}} 
  \caption{POS Tagging Accuracy Comparisons} 
  \label{posscatter} 
\end{figure}

Both the discrete and neural models get comparable accuracies with state-of-the-art system on English and Chinese datasets. The joint model has significant enhancement compared with separated model, especially in Chinese POS tagging task, with 1\% accuracy increment. Fig. \ref{posscatter} shows the accuracy comparison for the discrete and neural models based on each sentence. There are many sentences that are not located at the diagonal line, which indicates the two models gives different results and have the potential for combination. Our joint model outperforms state-of-the-art accuracy with 0.23\% and 0.97\% on English and Chinese datasets, respectively.
\begin{table}[!htbp]
\centering
\caption{POS Tagging Datasets Statistics}\label{tab:posdatasets}
\begin{tabular}{|c|c|c|}
\hline
POS tagging datasets & \multirow{2}*{English}& \multirow{2}*{Chinese}\\
(sentences) & &\\
\hline
Train& 38219& 16091\\
\hline
Dev& 5527& 803\\
\hline
Test& 5462& 1910\\
\hline
\end{tabular}
\end{table}

\begin{table}[!htbp]
\centering
\caption{POS Tagging Results}\label{tab:posresults}
\begin{tabular}{|c|c|c|}
\hline
 \multirow{2}*{Model}& English& Chinese\\
\cline{2-3}
 & Acc& Acc\\
\hline
Discrete& 97.23& 93.97\\
\hline
Neural& 97.28& 94.02\\
\hline
Joint& \textbf{97.47}& \textbf{95.07}\\
\hline
State-of-the-art& 97.24&  94.10\\
\hline
\end{tabular}
\end{table}

\subsection{NER}
For the NER task, we split Ontonotes 4.0 following Che et al. \cite{che2013named} to get both English and Chinese datasets. Table \ref{tab:nerdatasets} shows the sentence numbers of train/develop/test datasets.

\begin{table}[!t]
\centering
\caption{NER Datasets Statistics}\label{tab:nerdatasets}
\begin{tabular}{|c|c|c|}
\hline
NER datasets & \multirow{2}*{English}& \multirow{2}*{Chinese}\\
(sentences) & &\\
\hline
Train& 39262& 15724\\
\hline
Dev& 6249& 4301\\
\hline
Test& 6452& 4346\\
\hline
\end{tabular}
\end{table}

\begin{table}[!t]
\centering
\caption{NER Results}\label{tab:nerresults}
\begin{tabular}{|c|ccc|ccc|}
\hline
 \multirow{2}*{Model}&  \multicolumn{3}{|c|}{English}& \multicolumn{3}{|c|}{Chinese}\\
\cline{2-7}
 &  P& R& F& P& R& F\\
\hline
Discrete&  80.14& 79.29& 79.71& 72.67& 73.92& 73.29\\
\hline
Neural&  77.25& 80.19& 78.69& 65.59& 71.84& 68.57\\
\hline
Joint&  81.90& \textbf{83.26}& \textbf{82.57}& 72.98& \textbf{80.15}& \textbf{76.40}\\
\hline
State-of-the-art&  \textbf{81.94}& 78.35& 80.10& \textbf{77.71}& 72.51& 75.02\\
\hline
\end{tabular}
\end{table}

\begin{figure}[!t] 
  \centering 
  \subfigure[English]{ 
    \label{fig:neren} 
    \includegraphics[width=2.2in]{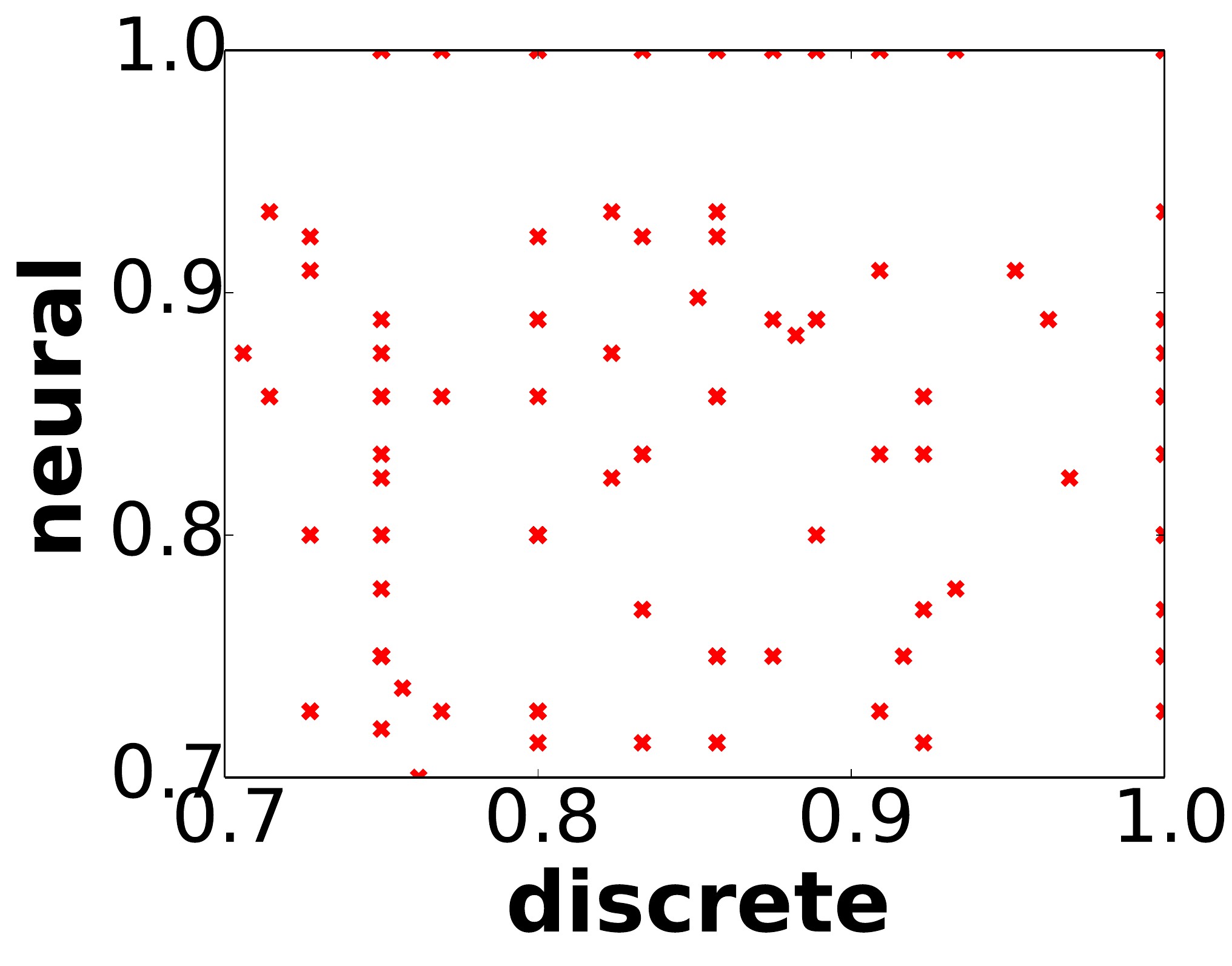}} 
  \subfigure[Chinese]{ 
    \label{fig:nercn} 
    \includegraphics[width=2.2in]{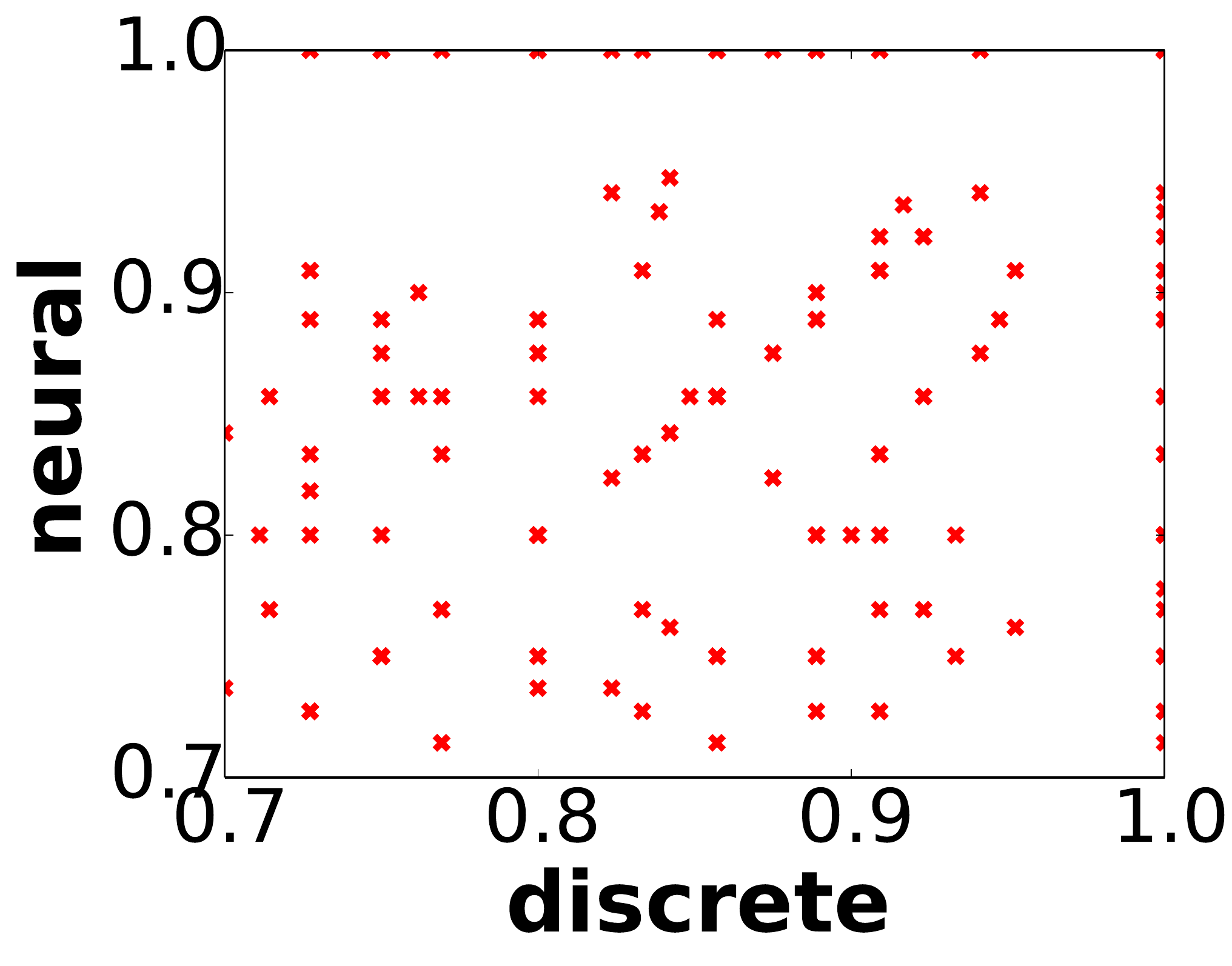}} 
  \caption{NER F-measure Comparisons} 
  \label{nerscatter} 
\end{figure}

We follow Che et al. \cite{che2013named} on choosing both the English and the Chinese datasets. Their work induces  bilingual constrains from parallel dataset which gives significant enhancement of F-scores on both English and Chinese datasets. 
 
 Our discrete and neural models show comparable recall values compared with Che's results \cite{che2013named} on both datasets. Similar with the previous two tasks, the joint model gives significant enhancement compared with separated models (discrete/neural) on all metrics. This shows that discrete and neural model can identify entities using different indicator features, and they can be complementary with each other. The comparison of sentence F-measures in Fig. \ref{nerscatter} confirms this observation.
 
 The joint model outperforms the state-of-the-art on both datasets in the two different languages. 
 The precision of joint models are less than state-of-the-art system. This may be caused by the bilingual constrains in baseline system, which ensures the precision of entity recognition.

\section{Conclusion}
We proposed a joint sequence labeling model that combines neural features and discrete indicator features which can integrate 
the advantages of carefully designed feature templates over decades and automatically induced features from neural networks. 
Through experiments on various sequence labeling tasks, including Chinese word segmentation, POS tagging and named entity recognition for Chinese and English respectively, we demonstrate that our joint model can unanimously outperform models which only contain discrete features or neural features and state-of-the-art systems on all compared tasks.The accuracy/F-measure distribution comparison for discrete and neural model also indicate that discrete and neural model can reveal different related information, this explains why combined model can outperform separate models.

In the future, we will investigate the effect of our joint model on more NLP tasks, such as parsing and machine translation.

\section*{Acknowledgments}
We would like to thank the anonymous reviewers for their detailed comments. This work is supported by the Singapore Ministry of Education (MOE) AcRF Tier 2 grant T2MOE201301.
\newpage

\bibliographystyle{unsrt}
\bibliography{references}

\end{document}